\documentclass[a4paper]{article}

\usepackage{fullpage} 
\usepackage{parskip} 
\usepackage{tikz} 
\usepackage{amsmath}
\usepackage{hyperref}
\usepackage{subfig}

\usepackage{graphicx}             
\usepackage{moreverb}             

\usepackage{graphicx}
\usepackage{amsmath,amssymb} 
\usepackage{color}
\usepackage{blindtext}
\usepackage{tabto, hyperref}
\usepackage{subfig, array}
\usepackage{blindtext, amsmath}
\usepackage{graphicx, tabto, hyperref}
\usepackage{subfig}
\usepackage{dblfloatfix}
\usepackage{graphics}
\usepackage{fontenc}
\usepackage{txfonts}
\usepackage{pifont}
\usepackage{color}
\usepackage{hyperref}
\usepackage{booktabs}
\usepackage{multirow}
\usepackage{float}
\usepackage{siunitx, array}
\usepackage{authblk}

\title{PProCRC: Probabilistic Collaboration of Image Patches}

\author[1]{Tapabrata Chakraborti}
\author[1]{Brendan McCane}
\author[1]{Steven Mills}
\author[2]{Umapada Pal}

\affil[1]{Dept. of Computer Science, University of Otago, NZ}
\affil[2]{CVPR Unit, Indian Statistical Institute, India}

\begin{document}

\maketitle

\begin{abstract}

We present a conditional probabilistic framework for collaborative representation of image patches. It incorporates background compensation and outlier patch suppression into the main formulation itself, thus doing away with the need for pre-processing steps to handle the same. A closed form non-iterative solution of the cost function is derived. The proposed method (PProCRC) outperforms earlier CRC formulations: patch based (PCRC, GP-CRC) as well as the state-of-the-art probabilistic (ProCRC and EProCRC) on three fine-grained species recognition datasets (Oxford Flowers, Oxford-IIIT Pets and CUB Birds) using two CNN backbones (Vgg-19 and ResNet-50).

\end{abstract}

\section{Introduction}
\label{sec:sec1}

Object recognition from images for categories with limited training samples or with fine-grained differences remains an open challenge [1]. In such problems, it is challenging to effectively train deep networks, even when fine-tuning a pre-trained base object classifier through established transfer learning methods [2]. Considering the case of fine-grained species recognition as a representative problem [3], besides scarcity of training data, there are further bottlenecks like subtle inter-class object differences compared to significant randomized background variation both between and within classes [4]. Added to these, is the presence of the ``long tail'' problem, that is, significant imbalance in samples per class (the frequency distribution of samples per class has long tail) [5].

Transfer learning is a popular approach in dealing with deep learning of small challenging datasets [2]. The ConvNet architecture is trained first on a large benchmark image dataset (eg. ImageNet [6]) for the task of base object recognition. The network is then fine-tuned on the target dataset for fine-grained recognition. If the number of samples per class is low, then the network cannot generalize to unknown test samples due to over-fitting. On the other hand, if the dataset has fine-grained objects with varying backgrounds, this can cause difficulty in achieving training convergence. This makes training on such datasets a challenge. In case of small datasets with imbalanced classes, the problem is compounded by the probability of training bias in favour of larger classes. So deep learning of small fine-grained datasets remains one of the open challenges of machine vision.

In our earlier work, we have demonstrated that collaborative filters can effectively represent and utilize small fine-grained datasets [7]. Collaborative filters are popular in recommender systems [8] to effectively encode user trends. Collaborative representation classification (CRC) represents the test image as an optimal weighted average of training images across all classes. The predicted label is the class having least residual. This inter-class collaboration for optimal feature representation is different compared to the traditional purely discriminative approach. CRC has a closed form solution and does not need iterative or heuristic optimization; thus it is efficient and analytic. It is also a general feature representation-classification scheme and thus most popular features and ensembles thereof are compatible with it. 


In computer vision, CRC was first applied to the face recognition problem by Zhang \emph{et al} [9]. This is because human faces have subtle inter-class differences and significant similarities across classes and CRC is effective in encoding these attributes across classes as mentioned before. However, most of the existing work on CRC based face recognition have reported results on benchmark datasets having well aligned and centered images with minimal background. Even the few works which have used face datasets in natural scene backgrounds have mostly employed pre-processing steps to align and crop the face region, thus removing the effect of the natural setting by manual intervention [10].

It has been shown that the performance of these methods degrades considerably when there is significant background which is randomised across classes [3]. This may be found in such fine-grained recognition problems like species recognition with varying habitats. Many variations of CRC have been proposed but most, if not all, carry this drawback. One approach for overcoming this is to use majority voting by patches, where the background effect gets compensated if it is randomly distributed across classes [11]. However, these methods still need to take into account several conditions like whether the test patch itself is an outlier, whether the patches predict the same label as the entire image, etc. 

The present work overcomes the above drawbacks of the existing methods.We present a new conditional probabilistic framework for collaborative representation of image patches (PProCRC) that handles outlier background patches better than its predecessors. Background suppression is formulated into the main cost function, thus doing away with the need for initial pre-processing steps like detection/localisation (annotation, bounding box, cropping). We present a closed form analytic solution of the cost function that is non-iterative and hence time efficient. The proposed method outperforms several competing methods including the state-of-the-art CRC methods on species recognition tasks.

\begin{figure*}[t!]
\centering

\subfloat{\includegraphics[width=3in,height=2.5in]{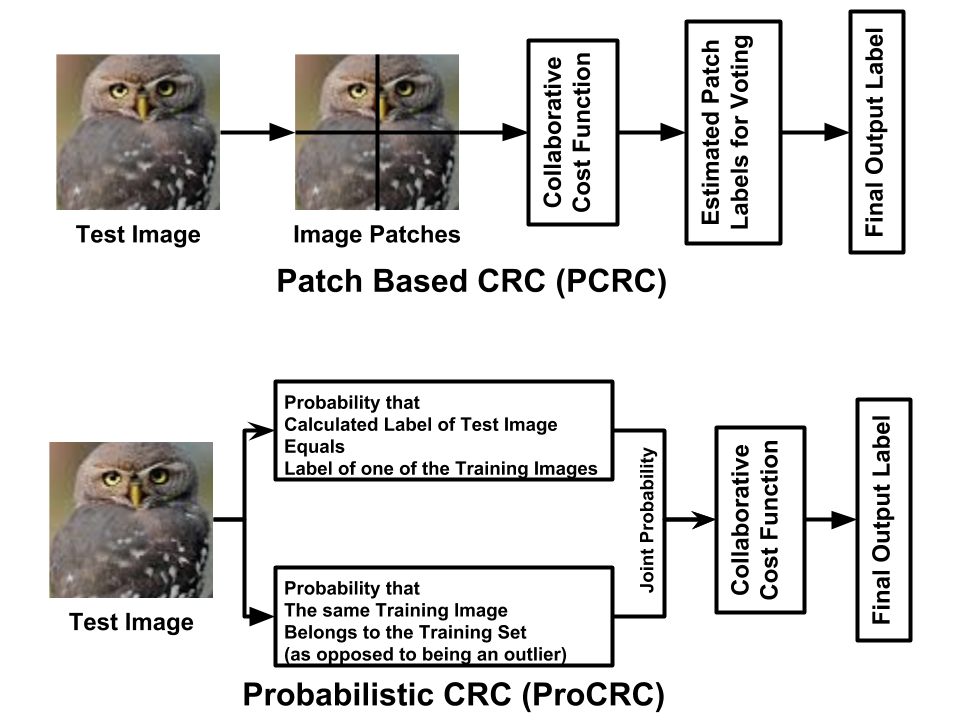}}
\hfil
\subfloat{\includegraphics[width=3in,height=2.5in]{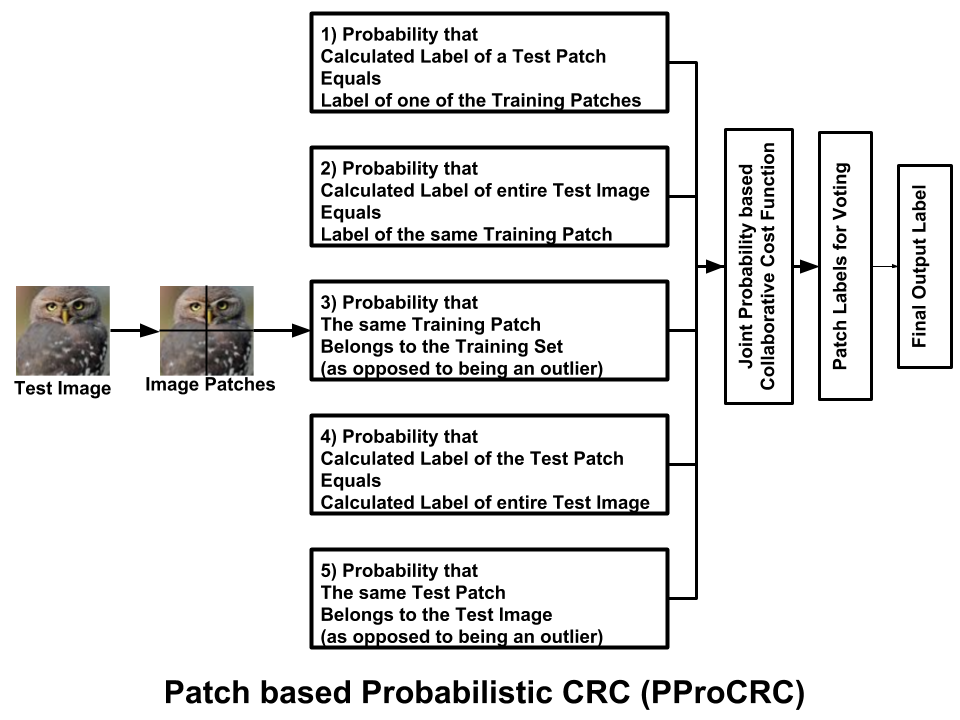}}
\caption{Schematic Diagram of the proposed method PProCRC and its direct predecessors PCRC and ProCRC.}
\label{fig:crc}
\end{figure*}

\section{Related Previous Collaborative Classifiers}
\label{sec:sec2}

In this section, we present briefly the original CRC and its two variants that the proposed method tries to improve. These are the patch based CRC (PCRC) and the probabilistic CRC (ProCRC) methods. Our probabilistic patch based CRC (PProCRC) tries to overcome the drawbacks of these existing methods.

\subsection{Collaborative Representation Classification (CRC)}

The mathematical framework for Collaborative Representation Classification (CRC) [9] is described in brief here. Consider a training set with images in some feature space as $X=[X_1,\dots,X_c]\in \varmathbb{R}^{d \times N}$ where $N$ is the total number of samples over $c$ classes and $d$ is the feature dimension per sample. Thus $X_i \in \varmathbb{R}^{d \times n_i}$ is the feature space representation of class $i$ with $n_i$ samples such that $\sum_{i=1}^{c} n_i = N$.

The CRC model reconstructs a test image in the feature space $y \in \varmathbb{R}^d$ as an optimal collaboration of all training samples, while at the same time limiting the size of the reconstruction parameters, using the regularization term $\lambda$.

The CRC cost function is given as 
\begin{equation}
J(\alpha,\lambda)=\text{arg}\,\min\limits_{\alpha}\,(\|y-X\alpha\|_2^2+\lambda\|\alpha\|_2^2)
\end{equation}                  
where $\hat{\alpha}=[\hat{\alpha}_1,\dots,\hat{\alpha}_c]\in\varmathbb{R}^N$ and $\hat{\alpha}_i\in\varmathbb{R}^{n_i}$ is the reconstruction matrix corresponding to class $i$.

\noindent A least-squares derivation yields the optimal solution as
\begin{equation}
\hat{\alpha}=(X^TX+ \lambda I)^{-1}X^Ty
\end{equation}     

\noindent The representation residual of class $i$ for test sample $y$ can be calculated as: 
\begin{equation}
r_i(y)=\frac{\|y-X_i\hat{\alpha}_i\|_2^2}{\|\hat{\alpha}_i\|_2^2} \ \forall i \in {1,\dots,c}
\end{equation}

\noindent The final class of test sample $y$ is thus given by
\begin{equation}
C(y)=\text{arg}\,\min\limits_i\, r_i(y)
\end{equation}  


\subsection{Patch based CRC (PCRC)}

Zhu \emph{et al}. [11] introduced a patch-based framework for collaborative representation (PCRC). Let the query image $y$ be divided into $q$ overlapping patches $y=\{y_1,\dots,y_q\}$. 
From the feature matrix $X$, a local feature matrix $M_j$ is extracted corresponding to location of patch $y_j$. Thus the modified cost function becomes: 

\begin{equation}
\hat{p}_j=\text{arg}\,\min\limits_{p_j}\,(\|y_j-M_jp_j\|_2^2+\lambda\|p_j\|_2^2)
\end{equation}  
where $M_j=[M_{j1},\dots,M_{jc}]$ are the local dictionaries for the $c$ classes and $\hat{p}_j=[\hat{p}_{j1},\dots,\hat{p}_{jc}]$ is the optimal reconstruction matrix for the patch $j$.
The class of a patch in the test image is predicted as:
\begin{equation}
C(y_j)=\text{arg}\,\min\limits_k\, r_{jk}(y)
\end{equation}
where 
\begin{equation}
r_{jk}=\frac{\|y_j-M_{jk}\hat{p}_{jk}\|_2^2}{\|\hat{p}_{jk}\|_2^2} \ \forall i \in {1,\dots,c}
\end{equation}

The classification of the entire test sample $y$ is determined by majority voting of the classification labels of the patches $y_j$.

\subsection{Probabilistic CR Classifier (ProCRC)}






Cai \emph{et al}. [10] presented a probabilistic formulation (ProCRC) where each of the terms are modeled by Gaussian distributions and the final cost function for ProCRC is formulated as maximisation of the joint probability of the test image belonging to each of the possible classes as independent events. The final classification is performed by checking which class has the maximum likelihood. 

\begin{equation}
J(\alpha,\lambda,\gamma)=\|y-X\alpha\|_2^2+\lambda\|\alpha\|_2^2+ \frac{\gamma}{K}\sum_{k=1}^{K}\|X\alpha-X_k\alpha_k\|_2^2
\end{equation}  





\subsection{Drawbacks of earlier formulations}

The proposed PProCRC method overcomes the drawbacks of the PCRC and ProCRC methods, on which it is based. ProCRC gives a logical probabilistic framework to the CRC formulation, but suffers from the same drawback of most collaborative formulations, that of randomized background variation across fine-grained classes. For example, in the case of sub-categorical species recognition, the collaborative filter produces a robust representation of the fine-grained classes, but these species classes often contain a wide range of background variation in habitat which may be repeated randomly across classes, thus acting as a confounding factor for the inter-class collaborative representation. 

PCRC and other patch based CRC methods tend to overcome the background challenge by having a majority voting based classification scheme as described before. This might compensate for the effect of background patches if they are in the minority or if the background patches are randomised across classes which is often the case. However, patch based methods are prone to outliers if some images have rare backgrounds. Also, for images taken in the wild, the acutal object may take up a much smaller space than the background, thus resulting in many patches actually not having the object of interest at all. Our patch based probabilistic formulation of collaborative representation overcomes these challenges as discussed in the following Section.

\section{Probabilistic Patch based CR Classifier (PProCRC)}
\label{sec:sec3}

We formulate the proposed PProCRC cost function as a maximisation of the joint occurrence of three independent events that overcome the drawbacks of the earlier methods, while preserving the strengths of each. The main insight is that the predicted label of a patch ($y_i$) and the entire test image ($Y$) should be the same (that is same label as that of one of the patches $x$ of the training set to which it is the most similar in the collaborative space). This should be achieved under the condition that the patch $y_i$ is not an outlier in the test image $Y$ and that the training patch $x$ also is not an outlier. An example of this can be a rare background patch which is not commonly repeated in the dataset, and hence is assigned low probability so as not to affect the voting outcome. $\alpha_i$ and $\beta_i$ are the reconstruction vectors in following equations.

These probabilities are modeled as Gaussians and separated into three independent events as follows.

\begin{enumerate}
    \item Probability of a test patch having same label as one of the training patches and that training patch is not an outlier is given by:
    \begin{equation}
        P[l(y_i)=l(x) \mid x \in X].\ P[x \in X]=e^{-\|y_i-X\alpha_i\|}e^{-\lambda\|\alpha_i\|}
    \end{equation}
    \item Probability of the test patch having the same label as the total test image and that the test patch is not an outlier in the test image is given by:
    \begin{equation}
        P[l(y_i)=l(y) \mid y \in Y].\ P[y \in Y]=e^{-\|y_i-Y\beta_i\|}e^{-\gamma \|\beta_i\|}
    \end{equation}
    \item Probability of the entire test image having the same label as the training patch (which has same label as test patch) is given by:
    \begin{equation}
        P[l(y)=l(x)]=e^{-\|Y\beta_i-X\alpha_i\|}
    \end{equation}
\end{enumerate}

So the final cost function is given by the maximum of the joint occurrence of these 3 events as: \\ \\
    $\max\limits_{\alpha_i,\beta_i}\,[\exp(-\|y_i-X\alpha_i\|-\|y_i-Y\beta_i\|-\lambda\|\alpha_i\|-\gamma \|\beta_i\|-\|Y\beta_i-X\alpha_i\|)]$ \\
    \begin{equation}
   =\min\limits_{\alpha_i,\beta_i}\,[-\|y_i-X\alpha_i\|-\|y_i-Y\beta_i\|-\lambda\|\alpha_i\|-\gamma \|\beta_i\|-\|Y\beta_i-X\alpha_i\|]
   \end{equation}
Next we obtain a closed form solution of the cost function.

\begin{itemize}
    \item Differentiating with respect to $\alpha_i$ we have:
    \begin{equation}
        (2X^{T}X+\lambda I)\hat{\alpha}_i - X^{T}Y\hat{\beta}_i = X^{T}y_i
    \end{equation}
    \item Differentiating with respect to $\beta_i$ we have:
        \begin{equation}
        (2Y^{T}Y+\gamma I)\hat{\beta}_i - Y^{T}X\hat{\alpha}_i = Y^{T}y_i
    \end{equation}
\end{itemize}

Solving the simultaneous equations 13 and 14, we get the optimal values of $\hat{\alpha}_i$ and $\hat{\beta}_i$ as follows:

    \begin{multline}
        \hat{\alpha}_i = [(X^{T}Y)^{-1}(2X^{T}X+\lambda I)) - (2Y^{T}Y+\gamma I))^{-1}Y^{T}X]^{-1} . \\ [2Y^{T}Y+\gamma I))^{-1}Y^{T} + (X^{T}Y)^{-1}X^T] . y_i
    \end{multline}

and 
        
\begin{multline}
        \hat{\beta}_i = [(Y^{T}X)^{-1}(2Y^{T}Y+\gamma I)) - (2X^{T}X+\lambda I))^{-1}X^{T}Y]^{-1} . \\ [2X^{T}X+\gamma I))^{-1}X^{T} + (Y^{T}X)^{-1}Y^T] . y_i
    \end{multline}

These optimal values are then used for the classification phase through patch majority voting as in the PCRC scheme.

\textbf{Main advantage of PProCRC:} The proposed method incorporates certain conditional probabilistic penalties into the collaborative cost function that counteracts background variation, without the need for additional pre-processing steps. As an example, among other considerations, it also assigns penalties if a test image patch is dissimilar to training patches as well as to other patches in the test image, which mitigates the effect of outlier patches.

\section{Experiments and Results}
\label{sec:sec4}

 \subsection{Benchmark Datasets}

The proposed method and its competitors have been evaluated on three benchmark fine-grained classification datasets

\textbullet \ \emph{Oxford Flowers dataset:} It has 8,189 images of 102 flowers, with at least 40 images per class [15]. It was developed by the Robotics Group at Oxford University. It is an expansion of the earlier dataset by the same group with 17 flower types with 80 images per class [16]. 

\textbullet \ \emph{Oxford-IIIT Pets dataset:} This dataset, compiled by the Oxford Robotics Group and IIIT Hyderabad, consists of 37 categories of pet cats and dogs with around 200 images belonging to each class [17].

\textbullet \ \emph{CUB Birds dataset:} dataset contains 11,788 images of 200 bird species. The main challenge of this dataset is considerable variation and confounding features in background information compared to subtle inter-class differences in birds.

\begin{table}[ht!]

\renewcommand{\arraystretch}{2}

\caption{{\textbf{Species Recognition Accuracy (\%)}}}
\label{tab:tab2}
\centering
{\small
\begin{tabular}{|l|p{1cm}p{1cm}|p{1cm}p{1cm}|p{1cm}p{1cm}|}

\toprule
    \multirow{2}{*}&
      \multicolumn{2}{p{2cm}|}{\centering{\textbf{{Flowers}}}} &
      \multicolumn{2}{p{2cm}|}{\centering{\textbf{{Pets}}}} \\
      & \textbf{ResNet} & \textbf{VGG} & \textbf{ResNet} & \textbf{VGG} \\
      \midrule \hline


\textbf{SVM} & 91.5 & 90.9 & 84.2 & 82.6  \\
\hline
\textbf{RDF} & 93.2 & 92.6 & 85.6 & 83.5  \\
\hline
\textbf{CRC} & 93.0 & 91.8 & 85.1 & 83.3  \\
\hline
\textbf{ECRC} & 93.6 & 93.2 & 86.5 & 84.9  \\
\hline
\textbf{PCRC} & 93.9 & 93.0 & 86.9 & 84.8  \\
\hline
\textbf{RCRC} & 94.4 & 93.9 & 87.4 & 85.5  \\
\hline
\textbf{KCRC} & 94.7 & 94.1 & 87.8 & 85.7 \\
\hline
\textbf{ProCRC} & 95.1 & 94.8 & 89.5 & 86.9  \\
\hline
\textbf{GP-CRC} & 95.8 & 95.4 & 90.0 & 87.4  \\
\hline
\textbf{EProCRC} & 96.5 & \textbf{97.5} & 91.3 & 88.0  \\
\hline
\textbf{PProCRC} & \textbf{97.7} & 96.1 & \textbf{92.9} & \textbf{89.3} \\
\hline
\bottomrule

\end{tabular}
}
\end{table}

\subsection{Competing Classifiers}

\textbf{Non-CRC classifiers.} The performance of the proposed PProCRC method is compared with that of several competing classifiers, both CRC based as well as non-CRC based. We choose three popular modern non-CRC classifiers, namely support vector machines (SVM) and random decision forests (RDF).

\begin{itemize}

\item \emph{Support Vector Machines:} Multiclass categorization is performed with the binary SVM [19] classifier with $\chi^2$ kernel in a one-versus-all fashion. the kernel parameter gamma and the regularization parameter C are tuned. 


\item \emph{Random Decision Forest:} RDF [21] is an ensemble of Decision Trees. Since individual decision trees are prone to over-fitting, bootstrap aggregated (bagged) Classification and Regression Trees (CART) are used to achieve better generalisation. For Random Forest, the main approach was to set the number of trees high (10000) and then run a series of values for maximum depth parameter in an array to find the optimal value.

\end{itemize}

\textbf{CRC based classifiers.} We first take the ones that are directly related to the formulation of the present method. As has been described in Section \ref{sec:sec2}, these are the original CRC, patch based CRC (PCRC), generalized patch based CRC (GP-CRC) and probabilistic CRC (ProCRC). Besides these we also have used several other recent variations of CRC like Enhanced CRC (ECRC), Relaxed CRC (RCRC), Kernel CRC (CRC), and the recent Extended Probabilistic CRC (EProCRC). These are described briefly below. \\



\begin{itemize}

\item \emph{Enhanced Collaborative Representation (ECRC):} Liu \emph{et al.} [22] enhanced the original CRC by incorporating the covariance matrix $R$ of the training samples into the cost function:
\begin{equation}
\hat{\alpha}=\text{arg}\,\min\limits_{\alpha}\,\bigg((y-X\alpha)^TR^{-1}(y-X\alpha)+\lambda\|\alpha\|_2^2\bigg)
\end{equation}    




\item \emph{Relaxed Collaborative Representation (RCRC):} Yang \emph{et al.} [23] developed a CRC method (RCRC) with relaxed constraints assigning adaptive weights to features for optimal contribution to final representation. The weights are adjusted such that the variance of representative features from mean is controlled, in order to make the representation more stable.

Thus in the RCRC formulation, the cost function of CRC gets modified to 
\begin{equation}
\hat{\alpha}=\text{arg}\,\min\limits_{\alpha,w}\,\bigg(\|y-X\alpha\|_2^2+\lambda\|\alpha\|_2^2 + \tau w \|\alpha-\bar{\alpha}\|_2^2\bigg)
\end{equation} 
where $\tau$ is a positive constant and $w$ is the weight vector such that $w=[w_1,\dots,w_c]\mid w_i\in\Re$ and $c$ is the number of classes. \\


\item \emph{Kernel Collaborative Representation (KCRC):} Zhao \emph{et al.} [24] introduced the kernel trick into the CRC framework. 
The cost function for KCRC becomes:
\begin{equation}
\hat{\alpha}=\text{arg}\,\min\limits_{\alpha}\,\|\alpha\|_{l_p} \text{ subj. to } \|\phi(y)-\Phi\alpha\|_{l_q} \leq \epsilon
\end{equation}    
Here the second term imposes the kernel condition in higher dimension.

\item \emph{Generalised Patch based CR Classifier (GP-CRC):} Chakraborti \emph{et al}. [3] recently proposed a generalised enhancement (GP-CRC) of the basic patch based CRC (PCRC). The original PCRC only compares patches at the same corresponding location between images, which is a major drawback since this assumes that the foreground object is well centred, aligned and covers most of the image, which would rarely be the case for natural scene object recognition. 

First, GP-CRC constructs an augmented M with features of all patches over all training images, and then uses majority voting for final classification. This solution handles the case of misaligned foreground objects, but raises the chances of the representation learning the background. To compensate, it is further compared to location matched patches in order to have a penalty if the query patch is too dissimilar to other patches at same location. 
\begin{equation}
\hat{p}_j=\text{arg}\,\min\limits_{p_j}\,(\|y_j-Mp_j\|_2^2+\lambda\|p_j\|_2^2+\gamma\|Mp_j-M_jp_{jj}\|_2^2)
\end{equation}

This is a balance, trading off misaligned foreground objects with the risk of learning the background. Details of the equation may be found in [3]. \\

\item \emph{Extended Probabilistic CRC (EProCRC):} Lan \emph{et al}. [25] recently extended the probabilistic CRC model by incorporating an additional prior information metric $\beta_c$ into the cost function that measures the distance $\|X-X_k\|$ between the centroid of the training set from the centroid of the individual classes. Thus the predicted class label for a test sample $y$ is given by (symbols having usual meaning):
\begin{equation}
\hat{\alpha}=\text{arg}\,\min\limits_{\alpha}\,\bigg(\|y-X\alpha\|_2^2+\lambda\|\alpha\|_2^2+ \frac{\gamma}{K}\sum_{k=1}^{K}\beta_c\|X\alpha-X_k\alpha_k\|_2^2\bigg)
\end{equation}  



\end{itemize}

\subsection{CNN Backbone Architectures} 

We have used 2 popular CNN architectures for feature learning: ResNet-50 and Vgg-19. But it should be noted that the proposed algorithm is general and is agnostic to feature choice. They are pre-trained on more than one million images from the ImageNet [6] dataset, and can classify up to 1000 object categories. Adam optimiser is used and initital learning rate is 0.001. We have here fine-tuned the pre-trained models on our target datasets following the training protocols of [2], for further details [2] may be referred.








\subsection{Results}


For each dataset, experiments are conducted with five fold cross validation. Percentage classification accuracies along with standard deviation are presented in Table 1 with the highest accuracy in each column highlighted in bold. Among the CRC-based methods, basic CRC has the least accuracy and then there is a consistent increase in the performance of the CRC variants. The proposed Probabilistic patch based CRC (PProCRC) comfortably outperforms all the competing CRC methods including the two that it is based on, that is the original patch based CRC (PCRC) and the probabilistic CRC (ProCRC). It also has marginal improvement in performance over the recent enhanced probabilistic CRC (EProCRC). Compared to the non-CRC methods, PProCRC has significantly better results than both SVM and RDF. These results are consistent across the 6 datasets and 2 CNN backbones (with ResNet having better performance than VggNet).

\section{Conclusion}
\label{sec:sec6}

We present a new conditional probabilistic framework for collaborative representation of image patches (PProCRC) that handles outlier background patches better than its predecessors. The proposed method has outperformed several competing collaborative representation classification (CRC) methods including the state-of-the-art, as well as a few popular non-CRC classifiers on fine-grained species recognition tasks.

\bibliographystyle{plain}
\bibliography{References}

\begin{thebibliography}{1}



 
\bibitem{Chai15}
Y.~Chai, ``Advances in Fine-grained Visual Categorization.'', \emph{University of Oxford}, 2015.

\bibitem{Simon15}
M. Simon and E. Rodner, ``Neural Activation Constellations: Unsupervised Part Model Discovery with Convolutional Networks'', \emph{In Proc. ICCV}, 2015.

\bibitem{rodner15}
E.~Rodner, M.~Simon, G.~Brehm, S.~Pietsch, J.-W.Wägele, and J.~Denzler, ``Fine-grained Recognition Datasets for Biodiversity Analysis'', \emph{In Proc. CVPR}, 2015.

\bibitem{Chakraborti17}
T. Chakraborti, B. McCane, S. Mills, and U. Pal, ``A Generalised Formulation for Collaborative Representation of Image Patches (GP-CRC)'', \emph{In Proc. BMVC}, 2017.

\bibitem{Horn17}
G. V. Horn and P. Perona, ``The Devil is in the Tails: Fine-grained Classification in the Wild'', arXiv:1709.01450 [cs.CV], 2017.

\bibitem{Russakovsky15}
O. Russakovsky, J. Deng, H. Su, J. Krause, S. Satheesh, S. Ma, Z. Huang, A. Karpathy, A. Khosla, M. Bernstein, A. C. Berg, and Li Fei-Fei, ``ImageNet Large Scale Visual Recognition Challenge,'' \emph{IJCV}, 115(3):211-252, 2015.

\bibitem{Chakraborti16}
T. Chakraborti, B. McCane, S. Mills, and U. Pal, ``Collaborative representation based fine-grained species recognition'', \emph{In Proc. IVCNZ}, 2016.

\bibitem{Schafer07}
J. B. Schafer, D. Frankowski, J. Herlocker and S. Sen,
``Collaborative Filtering Recommender Systems'',
\emph{The Adaptive Web, Lecture Notes in Computer Science, Springer}, vol. 4321, pp. 291-324, 2018.

\bibitem{Zhang2011}
L.~Zhang, M.~Yang, and X.~Feng, ``Sparse representation or collaborative representation: Which helps face recognition?'', \emph{In Proc. ICCV}, 2011.

\bibitem{Cai2016}
S.~Cai, L.~Zhang, W.~Zuo, and X.~Feng, ``A probabilistic collaborative representation based approach for pattern classification'', \emph{CVPR}, 2016. 

\bibitem{Zhu2012}
P.~Zhu, L.~Zhang, Q.~Hu, and Simon~C.K. Shiu, ``Multi-scale patch based collaborative representation for face recognition with margin distribution optimization'', \emph{ECCV}, 2012.

\bibitem{Martinez98}
A.M. Martinez and R. Benavente, ``The AR Face Database'', \emph{CVC Technical Report}, no.24, 1998.

\bibitem{Huang07}
G. B. Huang, M. Ramesh, T. Berg, and E. Learned-Miller, ``Labeled Faces in the Wild: A Database for Studying Face Recognition in Unconstrained Environments'', \emph{University of Massachusetts, Amherst, Technical Report}, pp. 07--49, 2007. 

\bibitem{Taigman09}
Y. Taigman, L. Wolf and T. Hassne, ``Multiple One-Shots for Utilizing Class Label Information'', \emph{In Proc. BMVC}, 2009.


\bibitem{Nilsback07}
M-E. Nilsback and A. Zisserman, ``Delving into the whorl of flower segmentation'', \emph{In Proc. BMVC}, 2007.


\bibitem{Nilsback08}
M-E. Nilsback and A. Zisserman, ``Automated flower classification over a large number of classes'', \emph{ In Proc. ICVGIP}, 2008.


\bibitem{Parkhi12}
O. M. Parkhi, A. Vedaldi, A. Zisserman and C. V. Jawahar, ``Cats and Dogs'', \emph{In Proc. CVPR}, 2012.

\bibitem{Chakraborti19}
T. Chakraborti, B. McCane, S. Mills, and U. Pal, ``CoCoNet: Collaborative ConvNet for deep transfer learning of fine-grained classes'', \emph{In Proc. IVCNZ}, 2020.

\bibitem{Cortes95}
C. Cortes and V. Vapnik, ``A Support Vector Networks'', \emph{Machine Learning}, vol. 20, no. 3, pp. 273--297, 1995.

\bibitem{Freund99}
Y. Freund and R. E. Schapire, ``A Short Introduction to Boosting'', \emph{Journal of Japanese Society for Artificial Intelligence}, vol. 14(5):771--780, 1999.

\bibitem{Ho95}
T. K. Ho, ``Random Decision Forests'', \emph{In Proc. ICDAR}, 1995.

\bibitem{Liu14}
Z. Liu, X. Zhao, T. Huang, J. Pu and Y. Si, ``Enhanced collaborative representation based classification'', \emph{In Proc. IEEE Intl. Conf. on Information and Automation (ICIA)}, 2014.

\bibitem{Yang12}
M.~Yang, L.~Zhang, D.~Zhang, and S.~Wang, ``Relaxed collaborative representation for pattern classification'', \emph{In Proc. CVPR}, 2012.

\bibitem{Zhao14}
J. Zhao, Y. Wang and B. Liu, ``Kernel collaborative representation for face recognition'', \emph{In Proc. Intl. Conf. on Signal Processing (ICSP)}, 2014.

\bibitem{Lan17}
R. Lan and Y. Zhou, ``An extended probabilistic collaborative representation based classifier for image classification'', \emph{In Proc. IEEE Intl. Conf. on Multimedia and Expo (ICME)}, 2017.

\bibitem{Lowe99}
D.G. Lowe, ``Object recognition from local scale-invariant features'', \emph{In Proc. ICCV}, 1999.

\bibitem{Oliva01}
A. Oliva and A. Torralba, ``Modeling the Shape of the Scene: A Holistic Representation of the Spatial Envelope'',
\emph{IJCV}, 42(3):145--175, 2001.

\bibitem{Dalal05}
N. Dalal and B. Triggs, ``Histograms of oriented gradients for human detection'', \emph{In Proc. CVPR}, 2005.



\bibitem{simonyan14}
K.~Simonyan and A.~Zisserman.
\newblock Very deep convolutional networks for large-scale image recognition.
\newblock {\em In Proc. ICLR}, 2014.



\bibitem{Khan15}
N. Y. Khan, B. McCane, and S. Mills, ``Better than SIFT?'', \emph{Machine Vision and Applications}, 26(6):819-836, 2015.












\end{thebibliography}

\end{document}